\documentclass{article} % For LaTeX2e
\usepackage{iclr2026_conference,times}

\usepackage{hyperref}
\usepackage{url}
\usepackage{graphicx}
\usepackage{booktabs}
\usepackage{amsmath}
\usepackage{xcolor}

\title{Embedding-Only Uplink for Onboard Retrieval Under Shift in Remote Sensing}

\author{Sangcheol Sim \\
Telepix \\
\texttt{sim2real@telepix.net}}

\iclrfinalcopy % Uncomment for camera-ready version, but NOT for submission.
\begin{document}
\maketitle
\footnotetext{Accepted at the Machine Learning for Remote Sensing (ML4RS) Workshop, ICLR 2026.}

\begin{abstract}
Downlink bottlenecks motivate onboard systems that prioritize hazards without transmitting raw pixels.
We study a strict setting where a ground station uplinks \emph{only} compact embeddings plus metadata, and an onboard system performs vector search to triage new captures.
We ask whether this embedding-only pipeline remains useful under explicit remote-sensing shift: \emph{cross-time} (pre/post-event), \emph{cross-event/location} (different disasters), \emph{cross-site cloud} (15 geographic sites), and \emph{cross-city AOI holdout} (buildings).
Using OlmoEarth embeddings on a scaled public multi-task benchmark (27 Sentinel-2 L2A scenes, 15 cloud sites, 5 SpaceNet-2 AOIs; 10 seeds), we find that all effective methods rely on the same uplinked embeddings, but the \emph{optimal decision head is task-dependent}: $k$NN retrieval is significantly superior for cloud classification ($0.92$ vs.\ centroid $0.91$; $p{<}0.01$, Wilcoxon), while class centroids dominate temporal change detection ($0.85$ vs.\ retrieval $0.48$; $p{<}0.01$).
These results show that embedding-only uplink is the key enabler---once embeddings are onboard, the system can select the best head per task at no additional uplink cost, with all telemetry under 1\,KB per query.
\end{abstract}

\section{Introduction}
Operational disaster response faces a mismatch between sensing capacity and limited contact windows for downlink \citep{denby2020orbital}.
Retrieval against a compact onboard memory can prioritize what to transmit; we focus on whether such retrieval remains effective under distribution shift for remote-sensing triage.

We study a strict setting: \textbf{ground-to-space uplink is embedding-only (plus metadata)}.
Hints are (embedding, metadata) tuples; the onboard system indexes them in a vector database and retrieves the top-$k$ most similar candidates ($k$ typically 1--10) for each new capture.
Rather than optimizing a single dataset, we ask whether this pipeline remains useful under \textbf{explicit RS shift axes} \citep{tuia2016domain}---\emph{cross-time} (pre vs.\ post disaster), \emph{cross-event/location} (different disasters and hard negatives), \emph{cross-site cloud} (15 geographic sites), and \emph{cross-city AOI holdout} for buildings---while emitting compact, auditable telemetry as the downlink product.

\textbf{Related work.}
ESA's $\Phi$-Sat-1 demonstrated onboard CNNs for EO, reducing downlink by ${\sim}90\%$ \citep{phisat1}; recent surveys cover foundation-model deployment on satellite platforms \citep{onboardsurvey2025}.
Self-supervised RS encoders---SatMAE \citep{satmae}, SpectralGPT \citep{spectralgpt}, OlmoEarth \citep{olmoearth}---yield capable representations; we evaluate whether they support retrieval under shift, not proposing a new encoder.
RAG \citep{lewis2020rag} augments LMs with retrieved context; we adapt retrieval to a satellite setting where ``generation'' is compact telemetry.

\begin{figure}[t]
  \centering
  \includegraphics[width=\linewidth]{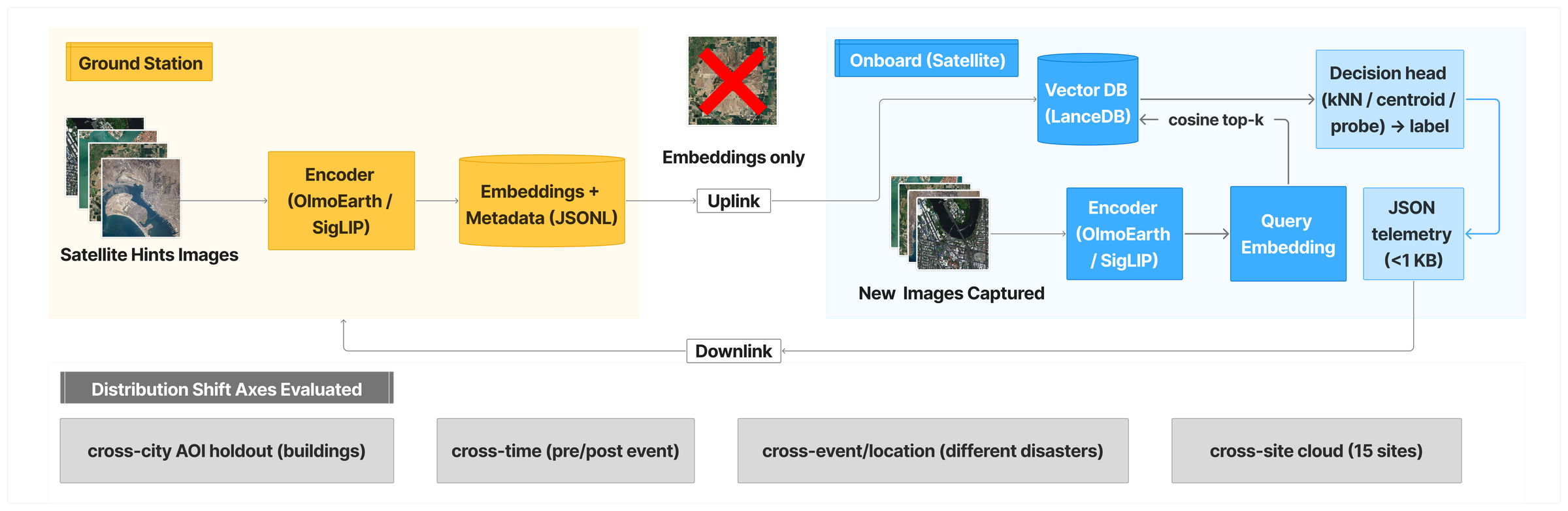}
  \caption{Embedding-only uplink pipeline. The ground station computes hint embeddings and uplinks (embedding, metadata) tuples---no imagery. The onboard system indexes hints in a vector database and retrieves top-$k$ matches for each new capture via cosine similarity, emitting compact JSON telemetry (${\sim}$700\,B at $k{=}5$) as the downlink product.}
  \label{fig:pipeline}
\end{figure}

\textbf{Contributions.} We do not propose a new encoder or retrieval algorithm; our contribution is the first systematic evaluation of embedding-only uplink for onboard RS triage under shift.
We contribute \textbf{(i)} an end-to-end pipeline for embedding-only uplink and onboard retrieval with compact telemetry (Figure~\ref{fig:pipeline}), \textbf{(ii)} a scaled reproducible multi-task benchmark (hazard, change, cloud, buildings; six baselines, 10 seeds, $k$-sweep, paired significance tests), and \textbf{(iii)} the finding that the optimal decision head is \emph{task-structure-dependent}, not difficulty-dependent: $k$NN retrieval excels for continuous-label tasks while class centroids dominate discrete temporal classification---but all heads rely on the same uplinked embeddings.\footnote{Code, data pipeline, and benchmark splits are released open-source: \url{https://github.com/weirdsim14/orbit-RAG}}

\section{Method}
\subsection{Embedding-only uplink and onboard indexing}
On the ground, we compute an embedding per hint image and store it with metadata in JSONL.
The onboard system ingests these rows into LanceDB \citep{lancedb} and performs cosine nearest-neighbor search at query time.
Crucially, uplink carries \emph{no imagery}.

\subsection{Multi-modal embeddings and decision head}
We use task-appropriate embedding backbones for public remote-sensing imagery:
OlmoEarth ($D{=}768$) for Sentinel-2 L2A 12-band patches \citep{olmoearth} and SigLIP ($D{=}1152$) for RGB building tiles \citep{siglip}.
L2A surface reflectance provides radiometrically consistent inputs; the 12-band spectral signature enables OlmoEarth to capture land-cover semantics without RGB reduction (single sensor: Sentinel-2 at 10\,m GSD for hazard/change/cloud; WorldView-3 via SpaceNet-2 for buildings).
For a query, we retrieve top-$k$ hints ($k\in\{1,5,10\}$) and apply a cosine-similarity-weighted $k$NN vote \citep{cover1967nearest} to produce a task label.
Patches are $256{\times}256$ pixels; embeddings are L2-normalized after OlmoEarth's computed normalization.
Cloud labels use STAC \texttt{eo:cloud\_cover} thresholds (clear$\leq$10\%, cloudy$\geq$20\%).
As non-retrieval baselines, we evaluate (i)~a nearest-centroid prototype classifier \citep{snell2017prototypical} and (ii)~a ridge-regression linear probe \citep{alain2017understanding} ($\ell_2{=}10^{-3}$), both operating on hint embeddings only.

\subsection{Compact, auditable telemetry}
The onboard system emits a minimal JSON record per query (task id, label, top-$k$ hints with scores); we report the serialized byte size.
Uplink cost is $N_{\text{hints}}\cdot D\cdot b$ bytes per hint-set refresh ($b\in\{4,2,1\}$ for FP32/FP16/INT8).
Downlink telemetry averages 598--690\,B per query ($k{=}5$); retrieval latency is ${\sim}$5\,ms (LanceDB cosine, single CPU thread).

\section{Benchmark and metrics}
\subsection{Multi-task public benchmark}
We evaluate four tasks with a unified retrieval interface and \textbf{explicit split units} aligned with RS leakage risks \citep{roberts2017cross}.
The benchmark comprises: 27 Sentinel-2 scenes (${\sim}$70 hints, ${\sim}$30 queries) for hazard; 8 cross-time pairs for change; 15 geographic sites with 5 clear/cloudy scenes each (${\sim}$300 hints, ${\sim}$150 queries) for cloud; and 5 SpaceNet-2 AOIs (up to 30 tiles/AOI, ${\sim}$40 queries) for buildings.
All data sources are public/no-auth \citep{sentinel2,earthsearch,spacenet2}.
We repeat over 10 random seeds; seeds deterministically choose held-out quadrants (Sentinel-2 tasks) and per-AOI tile subsamples (SpaceNet) to report mean$\pm$std.

\textbf{(1)~Hazard retrieval} (Sentinel-2 L2A \citep{sentinel2,earthsearch,olmoearth}): Hazard retrieval tests whether embeddings cluster scenes by disaster type well enough to triage new captures as wildfire, flood, or normal without examining imagery.
Each scene is labeled by disaster-type group (wildfire, flood, or normal); hints are multiple quadrant crops per scene; queries use a held-out quadrant (leave-one-crop-out).
Retrieval tests whether embeddings group scenes by disaster type: success requires correct-group hints in the top-$k$ (Recall@$k$) and correct top-1 (Top-1 accuracy).
The task is challenging because spectral signatures partially overlap across wildfire, flood, and normal scenes; hard negatives arise when different disasters share visual characteristics.
Optional normal-scene queries measure false-positive rate.
\textbf{(2)~Change (cross-time preference)} \citep{shi2020change}: a query from the after scene is correct if similarity to the after group exceeds the before group, testing whether embeddings preserve temporal ordering for triage prioritization.
\textbf{(3)~Cloud classification} (Sentinel-2 L2A): labels from STAC \texttt{eo:cloud\_cover} threshold; site-holdout split with held-out quadrant crops (anti-leakage).
\textbf{(4)~Buildings presence} (SpaceNet-2 \citep{spacenet2}, 0 vs.\ 1+): hints from non-holdout AOIs, queries from a held-out AOI (cross-city); GeoTIFFs converted to 8-bit RGB with 2--98\% percentile stretch before SigLIP embedding.

\subsection{Metrics}
Per task: \textbf{Recall@$k$} (any of $k$ hints from correct group), \textbf{Top-1 accuracy} (nearest hint correct), \textbf{time-preference accuracy} (after-group sim $>$ before-group; tests temporal ordering for change triage), \textbf{balanced accuracy} (mean per-class recall; handles site-level class imbalance for cloud), \textbf{macro-F1} (mean per-class F1; handles label skew for buildings), and \textbf{payload size} (serialized JSON bytes per query).

\begin{figure}[t]
  \centering
  \includegraphics[width=\linewidth]{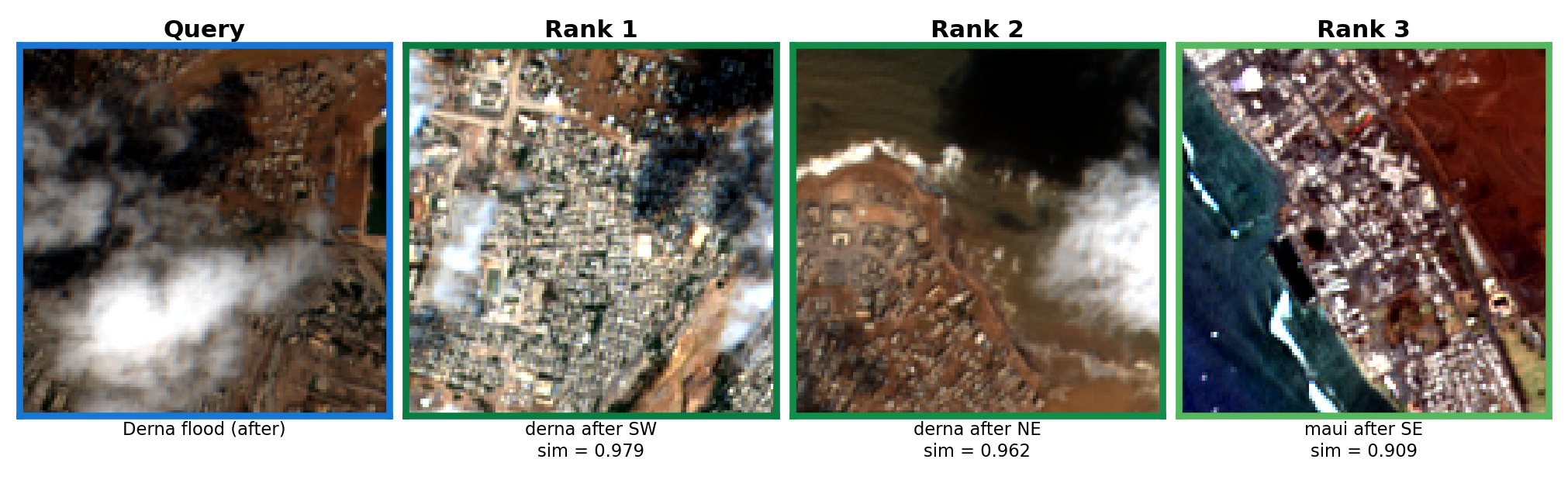}
  \caption{Qualitative retrieval example (hazard task). A Derna flood query retrieves same-event crops (sim$>$0.96) before a cross-event match (Maui wildfire, sim$=$0.91), demonstrating embedding-level hazard grouping.}
  \label{fig:qual}
\end{figure}

\section{Results}
All embedding-based methods significantly outperform embedding-free baselines, confirming that the uplinked embeddings are the key enabler.
However, the optimal decision head varies by task structure (Table~\ref{tab:main}; $k{=}5$, mean$\pm$std, 10 seeds; significance via paired Wilcoxon signed-rank test).

\textbf{Embeddings are the key enabler.}
Every embedding-based method (retrieval, centroid, linear probe) significantly outperforms random and no-retrieval baselines across all tasks ($p{<}0.01$), confirming that the uplinked embeddings---not the specific decision head---drive triage performance.
All telemetry remains under 700\,B per query ($k{=}5$).

\textbf{Optimal head is task-dependent.}
For \emph{cloud} (15 sites, ${\sim}$150 queries), $k$NN retrieval ($0.92$) significantly outperforms centroid ($0.91$; $p{=}0.004$).
Cloud cover is a continuous spectrum; $k$NN captures local similarity structure that centroid averaging smooths away.
For \emph{change} (8 cross-time pairs), centroid ($0.85$) significantly outperforms retrieval ($0.48$; $p{=}0.002$).
Before/after discrimination is a discrete class-level concept; the centroid captures the mean ``post-disaster'' pattern more robustly than nearest-neighbor matching.
For \emph{hazard}, retrieval and centroid both achieve near-perfect Recall@5 ($1.00$); Top-1 shows a non-significant advantage for retrieval ($0.91$ vs.\ $0.86$; $p{=}0.22$).
Figure~\ref{fig:qual} illustrates embedding-level hazard grouping qualitatively.
For \emph{buildings} under cross-city shift, centroid ($0.70$) trends above retrieval ($0.51$; $p{=}0.16$), but the difference is not statistically significant, likely due to limited AOI count (5) amplifying variance under cross-city shift.

\begin{figure}[t]
  \centering
  \includegraphics[width=\linewidth]{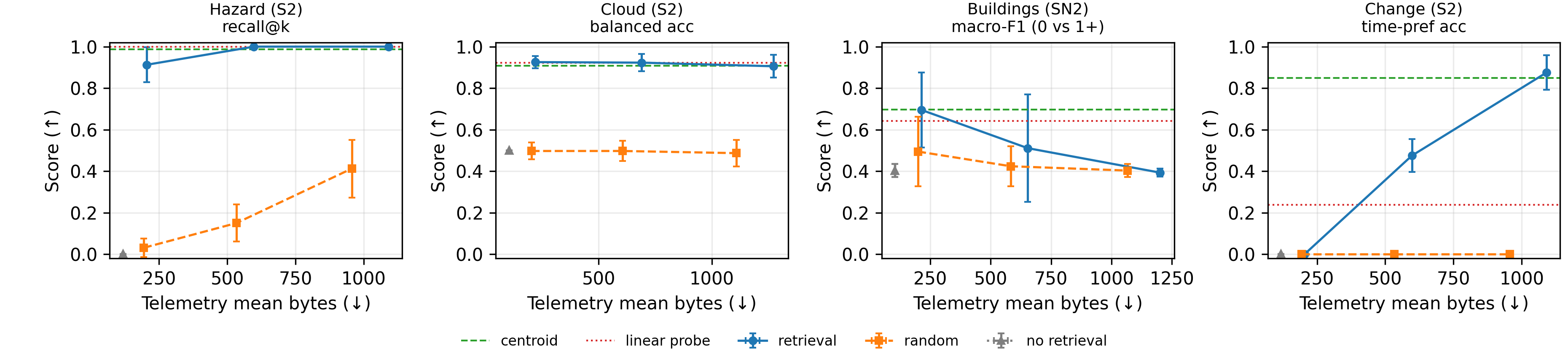}
  \caption{$k$-sweep: task metric vs.\ telemetry bytes ($k\in\{1,5,10\}$). Dashed horizontal lines show $k$-independent baselines (centroid, linear probe). Buildings favors small $k$; change improves with larger $k$. Error bars: $\pm$1\,std over 10 seeds.}
  \label{fig:ksweep}
\end{figure}

\textbf{$k$-sensitivity reveals task-specific operating points} (Figure~\ref{fig:ksweep}): buildings favors $k{=}1$ while change improves with larger $k$, motivating per-task selection in heterogeneous deployments.

\begin{table}[t]
  \caption{Multi-task results at $k{=}5$ (mean$\pm$std, 10 seeds, xlarge benchmark). Bold: best non-oracle per row. Significance markers from paired Wilcoxon signed-rank test vs.\ retrieval: $^{**}p{<}0.01$, $^{*}p{<}0.05$.}
  \label{tab:main}
  \centering\small
  \begin{tabular}{l l c c c c c c}
    \toprule
    Task & Metric & Retrieval & Random & Centroid & Lin.\ probe & No-retr. & Oracle \\
    \midrule
    Hazard (S2) & Recall@5 & $\mathbf{1.00}{\scriptstyle\pm.00}$ & $.15{\scriptstyle\pm.09}^{**}$ & $.99{\scriptstyle\pm.04}$ & $\mathbf{1.00}{\scriptstyle\pm.00}$ & $.00{\scriptstyle\pm.00}^{**}$ & $1.00{\scriptstyle\pm.00}$ \\
    Hazard (S2) & Top-1 & $\mathbf{.91}{\scriptstyle\pm.08}$ & $.03{\scriptstyle\pm.03}^{**}$ & $.86{\scriptstyle\pm.12}$ & $.90{\scriptstyle\pm.08}$ & $.00{\scriptstyle\pm.00}^{**}$ & $1.00{\scriptstyle\pm.00}$ \\
    Change (S2) & Time-pref & $.48{\scriptstyle\pm.08}$ & $.00{\scriptstyle\pm.00}^{**}$ & $\mathbf{.85}{\scriptstyle\pm.15}^{**}$ & $.24{\scriptstyle\pm.19}^{**}$ & $.00{\scriptstyle\pm.00}^{**}$ & $1.00{\scriptstyle\pm.00}$ \\
    Cloud (S2) & Bal.\ acc & $\mathbf{.92}{\scriptstyle\pm.04}$ & $.50{\scriptstyle\pm.05}^{**}$ & $.91{\scriptstyle\pm.05}^{**}$ & $.92{\scriptstyle\pm.04}$ & $.50{\scriptstyle\pm.00}^{**}$ & $1.00{\scriptstyle\pm.00}$ \\
    Build.\ (SN2) & Macro-F1 & $.51{\scriptstyle\pm.26}$ & $.42{\scriptstyle\pm.10}$ & $\mathbf{.70}{\scriptstyle\pm.12}$ & $.64{\scriptstyle\pm.17}$ & $.40{\scriptstyle\pm.03}$ & $1.00{\scriptstyle\pm.00}$ \\
    \bottomrule
  \end{tabular}
\end{table}

\section{Conclusion}
Embedding-only uplink supports multi-task RS triage under shift with ${\sim}$700\,B telemetry ($k{=}5$): $k$NN retrieval excels for continuous-label tasks (cloud, $p{<}0.01$) while centroids dominate discrete temporal classification (change, $p{<}0.01$). All heads share the \emph{same} uplinked embeddings, so practitioners select heads per task onboard at no additional uplink cost. Task structure (continuous vs.\ discrete label) predicts the optimal head, enabling zero-shot head selection without held-out validation data.

\textbf{Limitations.} Single optical sensor (Sentinel-2 L2A, 10\,m GSD); cross-sensor, cross-season, noise-robustness, and score calibration remain future work. Near-perfect hazard Recall@5 on 27 scenes warrants larger-scale replication.

\bibliographystyle{iclr2026_conference}
\bibliography{references}

\end{document}